\documentclass{article}

\usepackage{microtype}
\usepackage{graphicx}
\usepackage{subfigure}
\usepackage{booktabs} 
\usepackage{afterpage}
\usepackage{enotez}
\let\footnote=\endnote

\usepackage{hyperref}

\usepackage[accepted]{icml2023}

\usepackage{amsmath}
\usepackage{amssymb}
\usepackage{mathtools}
\usepackage{amsthm}

\usepackage[capitalize,noabbrev]{cleveref}

\theoremstyle{plain}
\newtheorem{theorem}{Theorem}[section]
\newtheorem{proposition}[theorem]{Proposition}
\newtheorem{lemma}[theorem]{Lemma}
\newtheorem{corollary}[theorem]{Corollary}
\theoremstyle{definition}
\newtheorem{definition}[theorem]{Definition}
\newtheorem{assumption}[theorem]{Assumption}
\theoremstyle{remark}
\newtheorem{remark}[theorem]{Remark}

\usepackage[textsize=tiny]{todonotes}
\usepackage{rotating}

\icmltitlerunning{ICML 2023 Topological Deep Learning Challenge}

\begin{document}

\twocolumn[
\icmltitle{ICML 2023 Topological Deep Learning Challenge: \\ Design and Results}



\icmlsetsymbol{equal}{*}

\begin{icmlauthorlist}
\icmlauthor{Mathilde Papillon}{equal,organizer}
\icmlauthor{Mustafa Hajij}{equal,organizer,reviewer}
\icmlauthor{Helen Jenne}{reviewer,contributor}
\icmlauthor{Johan Mathe}{reviewer,contributor}
\icmlauthor{Audun Myers}{reviewer,contributor}
\icmlauthor{Theodore Papamarkou}{reviewer,contributor}
\icmlauthor{Tolga Birdal}{contributor}
\icmlauthor{Tamal Dey}{contributor}
\icmlauthor{Tim Doster}{contributor}
\icmlauthor{Tegan Emerson}{contributor}
\icmlauthor{Gurusankar Gopalakrishnan}{contributor}
\icmlauthor{Devendra Govil}{contributor}
\icmlauthor{Aldo Guzmán-Sáenz}{contributor}
\icmlauthor{Henry Kvinge}{contributor}
\icmlauthor{Neal Livesay}{contributor}
\icmlauthor{Soham Mukherjee}{contributor}
\icmlauthor{Shreyas N. Samaga}{contributor}
\icmlauthor{Karthikeyan Natesan Ramamurthy}{contributor}
\icmlauthor{Maneel Reddy Karri}{contributor}
\icmlauthor{Paul Rosen}{contributor}
\icmlauthor{Sophia Sanborn}{contributor}
\icmlauthor{Robin Walters}{contributor}
\icmlauthor{Jens Agerberg}{participant}
\icmlauthor{Sadrodin Barikbin}{participant}
\icmlauthor{Claudio Battiloro}{participant}
\icmlauthor{Gleb Bazhenov}{participant}
\icmlauthor{Guillermo Bernardez}{participant}
\icmlauthor{Aiden Brent}{participant}
\icmlauthor{Sergio Escalera}{participant}
\icmlauthor{Simone Fiorellino}{participant}
\icmlauthor{Dmitrii Gavrilev}{participant}
\icmlauthor{Mohammed Hassanin}{participant}
\icmlauthor{Paul Häusner}{participant}
\icmlauthor{Odin Hoff Gardaa}{participant}
\icmlauthor{Abdelwahed Khamis}{participant}
\icmlauthor{Manuel Lecha}{participant}
\icmlauthor{German Magai}{participant}
\icmlauthor{Tatiana Malygina}{participant}
\icmlauthor{Rubén Ballester}{participant}
\icmlauthor{Kalyan Nadimpalli}{participant}
\icmlauthor{Alexander Nikitin}{participant}
\icmlauthor{Abraham Rabinowitz}{participant}
\icmlauthor{Alessandro Salatiello}{participant}
\icmlauthor{Simone Scardapane}{participant}
\icmlauthor{Luca Scofano}{participant}
\icmlauthor{Suraj Singh}{participant}
\icmlauthor{Jens Sjölund}{participant}
\icmlauthor{Pavel Snopov}{participant}
\icmlauthor{Indro Spinelli}{participant}
\icmlauthor{Lev Telyatnikov}{participant}
\icmlauthor{Lucia Testa}{participant}
\icmlauthor{Maosheng Yang}{participant}
\icmlauthor{Yixiao Yue}{participant}
\icmlauthor{Olga Zaghen}{participant}
\icmlauthor{Ali Zia}{participant}
\icmlauthor{Nina Miolane}{organizer}
\end{icmlauthorlist}

\icmlaffiliation{organizer}{Challenge Organizer}
\icmlaffiliation{reviewer}{Challenge Reviewer}
\icmlaffiliation{contributor}{Challenge Contributor}
\icmlaffiliation{participant}{Challenge Participant}

\icmlcorrespondingauthor{Mathilde Papillon}{papillon@ucsb.edu}

\icmlkeywords{Topology, Deep Learning, open-source}

\vskip 0.3in
]

\printAffiliationsAndNotice{\icmlEqualContribution}




\begin{abstract}
This paper presents the computational challenge on topological deep learning that was hosted within the ICML 2023 Workshop on Topology and Geometry in Machine Learning. The competition asked participants to provide open-source implementations of topological neural networks from the literature by contributing to the python packages \texttt{TopoNetX} (data processing) and \texttt{TopoModelX} (deep learning). The challenge attracted twenty-eight qualifying submissions in its two-month duration. This paper describes the design of the challenge and summarizes its main findings. \textbf{Code:} \url{https://github.com/pyt-team/TopoModelX}. \textbf{DOI:} 10.5281/zenodo.7958513.
\end{abstract}

\section{Introduction}

\begin{figure}[ht]
	\centering
 	\includegraphics[width=\linewidth]{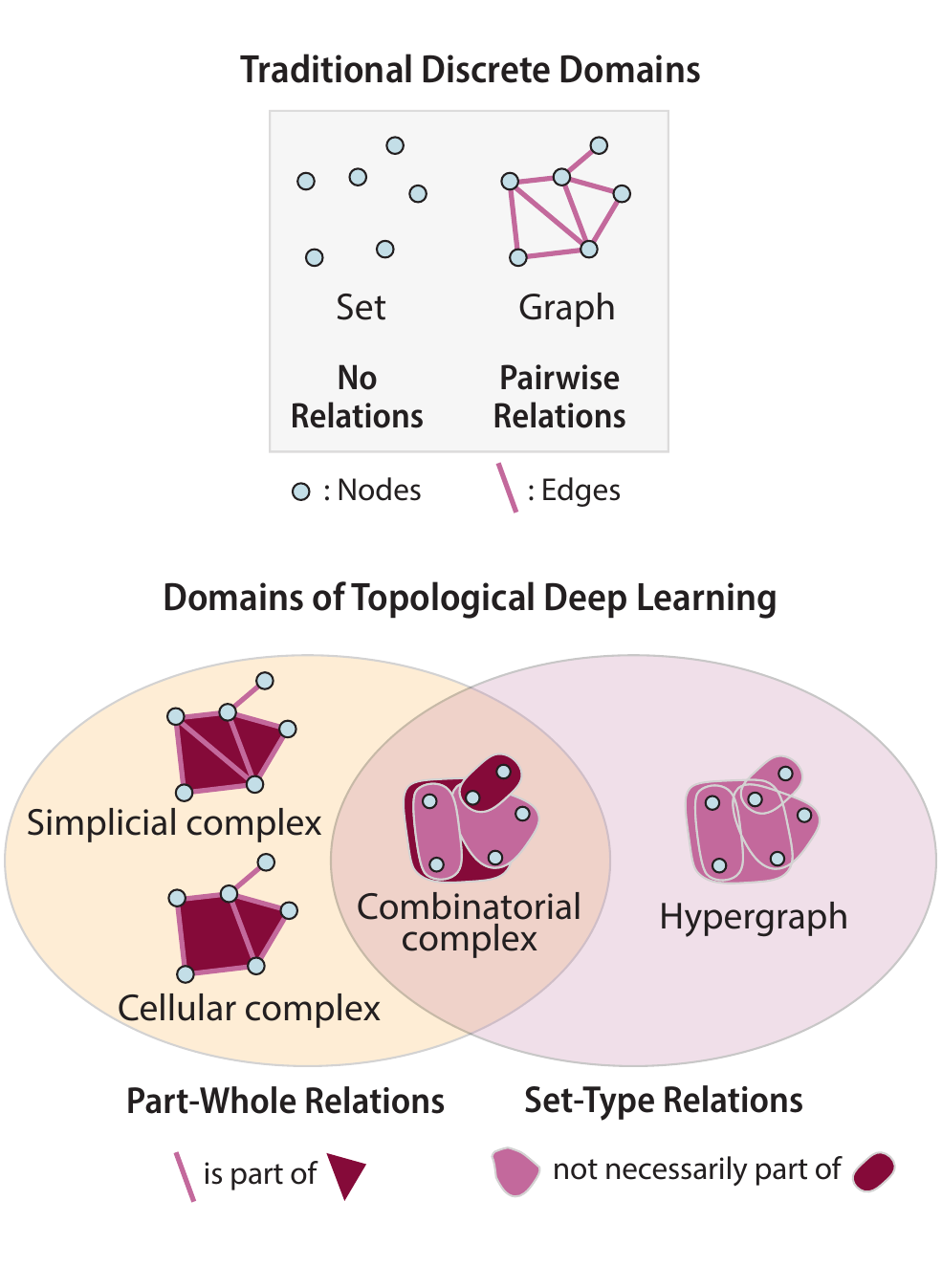}
	\caption{\textbf{Domains:} Nodes in light blue, (hyper)edges in pink, and faces in dark red. Adapted from \cite{hajij2023tdl}.}
	\label{fig:domains}
\end{figure}

Graph neural networks (GNNs) have proven to be a powerful deep learning architecture for processing relational data. More specifically, GNNs operate in graph domains comprised of pairwise relations between nodes. \textit{Topological neural networks} (TNNs) extend GNNs by operating on domains featuring higher-order relations. Such domains, called \textit{topological domains}, feature part-whole and/or set-type relations (Fig. \ref{fig:domains}) \cite{hajij2023tdl}, allowing a more expressive representation of the data. By operating on a topological domain, a TNN leverages the intricate relational structure at the heart of the data. Topological deep learning \cite{bodnar2023thesis,hajij2023tdl} has shown great promise in many applications, ranging from molecular classification  to social network prediction. However, the adoption of its architectures has been limited by the fragmented availability of open-source algorithms and lack of benchmarking between topological domains. 

The challenge described in this white paper aims to fill that gap by implementing models in a unifying open-source software. In doing so, the challenge contributes to fostering reproducible research in topological deep learning. Participants were asked to contribute code for a published TNN, following \texttt{TopoModelX}'s API \cite{hajij2023tdl} and computational primitives, and implement a training mechanism for the algorithm's intended task.

This white paper is organized as follows. Section~\ref{sec:setup} describes the setup of the challenge, including its guidelines and evaluation criteria. Section~\ref{sec:results} lists all qualifying submissions to the challenge and its winners.

\section{Setup of the challenge}\label{sec:setup}

The challenge \footnote{Challenge website: \url{https://pyt-team.github.io/topomodelx/challenge/index.html}} was held in conjunction with the workshop Topology and Geometry in Machine Learning of the International Conference on Machine Learning (ICML) 2023 \footnote{Topology and Geometry in Machine Learning Workshop website: \url{https://www.tagds.com/events/conference-workshops/tag-ml23}}. Participants were asked to contribute code for a previously existing TNN and train it on a toy dataset of their choice.

\paragraph{Guidelines} 

Each submission took the form of an implementation of a pre-existing TNN listed in a survey of the field \cite{papillon2023architectures}. These models fall into four categories, defined by their topological domain. All submitted code was required to comply with \texttt{TopoModelX}’s GitHub Action workflow~\cite{hajij2023tdl}, successfully passing all tests, linting, and formatting.

Each submission consisted of a pull request to \texttt{TopoModelX} containing three new files:

\begin{enumerate}
    \item A Python script implementing a layer of the model in a single class using \texttt{TopoModelX} computational primitives. One layer is equivalent to the message passing depicted in the tensor diagram representation for the model given in the survey~\cite{papillon2023architectures}.
    \item A Jupyter notebook that builds a neural network out of the single layer, loads and pre-processes the chosen dataset, and performs a train-test loop on the dataset. Defining training and testing in a Jupyter notebook offers authors a natural way to communicate results that are reproducible, as anyone with access to the notebook may run it to attain analogous results.
    \item A Python script which contains the unit tests for all methods stored in the class defining the model layer.
\end{enumerate}

Teams were registered to the challenge upon submission of their pull request and there was no restriction on the number of team members, nor on the amount of submissions per team. 

The principal developers of \texttt{TopoModelX} were not allowed to participate. Consistent with the aims of an open environment for sharing participation in this activity is completely voluntary and no support or endorsement of any of the participating parties by any of the other participating parties is provided. All submissions are the views of the individual participants only and should be taken, as is with all faults and without any guarantee, promise or endorsement of any kind.

\paragraph{Evaluation criteria} The evaluation criteria were:
\begin{enumerate}
\item Does the submission implement the chosen model correctly, specifically in terms of its message passing scheme? (The training schemes do not need to match that of the original model).
\item How readable and clean is the code? How well does the submission respect \texttt{TopoModelX}’s APIs?
\item Is the submission well-written? Do the docstrings clearly explain the methods? Are the unit tests robust?
\end{enumerate}
Note that these criteria were not designed to reward model performance, nor complexity of training. Rather, these criteria aimed to reward clean code and accurate model architectures that will foster reproducible research in topological deep learning.

\paragraph{Evaluation Method} The Condorcet method \cite{young1988condorcet} was used to rank the submissions and decide on the winners. Each team whose submission respected the guidelines was given one vote in the decision process. Nine additional reviewers selected from PyT-team maintainers and collaborators were also each given a vote. Upon voting, participating teams and reviewers were each asked to select the best and second best model implementation in each topological domain, thus making eight choices in total. Participants were not allowed to vote for their own submissions.

\paragraph{Software engineering practices} Challenge participants were encouraged to use software engineering best practices. All code had to be compatible with Python 3.10 and a reasonable effort had to be made for the code to adhere to PEP8 Python style guidelines. The chosen dataset had to be loaded from \texttt{TopoNetX}~\cite{hajij2023tdl} or \texttt{PyTorch-Geometric} \cite{pytorchgeo}. Participants could raise GitHub issues and/or request help at any time by contacting the organizers.

\section{Submissions and Winners}\label{sec:results}

In total, the challenge received 32 submissions, 28 of which adhered to the above outlined qualification requirements. Out of the qualifying submissions, 23 unique models were implemented. All four topological domains are represented in this set of models: 12 hypergraph implementations, 11 simplicial model implementations, 3 cellular implementations, and 2 combinatorial implementations. 


Table~\ref{submissions-table} lists all qualifying submissions. \cite{papillon2023architectures} contains additional information on the architectures and message-passing frameworks for each of these models.

Table~\ref{submissions-table} also indicates the winning contributions, consisting of a first and second prize for each topological domain, as well as honorable mentions. The winners were announced publicly at the ICML Workshop on Topology, Algebra and Geometry in Machine Learning and on social medias. Regardless of this final ranking, we would like to stress that all the submissions were of very high quality. We warmly congratulate all participants.

\section{Conclusion}

This white paper presented the motivation and outcomes of the organization of the Topological Deep Learning Challenge hosted through the ICML 2023 workshop on Topology, Algebra and Geometry in Machine Learning. Challenge submissions implemented a wide variety of topological neural networks into the open-source package \texttt{TopoModelX}. We hope that this community effort will foster reproducible research and further methodological benchmarks in the growing field of topological deep learning.

\section*{Acknowledgments}

The authors would like the thank the organizers of the ICML 2023 Topology, Algebra and Geometry in Machine Learning Workshop for their valuable support in the organization of the challenge.

\clearpage

\begin{table}[p]

\centering

\footnotesize

\begin{tabular}{lllllll}

\textbf{Domain} &
  \textbf{Model} &
  \multicolumn{3}{c}{\textbf{Task Level}} &
  \textbf{Computational challenge submission authors}\\ \midrule

 &
   & \rule{0pt}{33pt} 
  \begin{rotate}{90}\textit{Node} \end{rotate}&
  \begin{rotate}{90}\textit{Edge} \end{rotate} &
  \begin{rotate}{90}\textit{Complex} \end{rotate}&
   &
   \\ \toprule
\textbf{HG}&
  HyperSage \cite{arya2020hypersage} &
  \checkmark &
   &
   &
  \begin{tabular}[c]{@{}l@{}}German Magai, Pavel Snopov\end{tabular}  \\
  \cmidrule(lr){2-7}
 &
  AllSetTransformer \cite{chien2022you} &
  \checkmark &
   &
   &
   Luca Scofano, Indro Spinelli, Simone Scardapane, Simone \\
   & & & & & Fiorellino, Olga Zaghen, Lev Telyatnikov, Claudio \\
   & & & & & Battiloro, Guillermo Bernardez \textcolor{blue}{(first place)} \\
  \cmidrule(lr){2-7}

 &
   AllSetTransformer \cite{chien2022you} &
  \checkmark &
   &
   &
   Luca Scofano, Indro Spinelli, Simone Scardapane, Simone \\
   & & & & & Fiorellino, Olga Zaghen, Lev Telyatnikov, Claudio \\
   & & & & & Battiloro, Guillermo Bernardez \\
  \cmidrule(lr){2-7}

 &
  HyperGat \cite{ding2020less} &
  \checkmark &
   &
   &
  German Magai, Pavel Snopov \\
  \cmidrule(lr){2-7}

 &
  HNHN \cite{dong2020hnhn} &
  \checkmark &
  \checkmark &
   &
 1. Alessandro Salatiello \textcolor{orange}{(hon. mention)}\\
 & & & & & 2. Sadrodin Barikbin
   & \\
    \cmidrule(lr){2-7}

 &
  HMPNN* \cite{heydari2022message} &
  \checkmark &
   &
   &
  Sadrodin Barikbin \textcolor{red}{(second place)}\\
    \cmidrule(lr){2-7}

 &
  UniGCN \cite{huang2021unignn} &
  \checkmark &
   &
   &
  Alexander Nikitin \textcolor{orange}{(hon. mention)}\\
    \cmidrule(lr){2-7}
 &
  UniSAGE \cite{huang2021unignn} &
  \checkmark &
   &
   &
  Alexander Nikitin \\
    \cmidrule(lr){2-7}
 &
  UniGCNII \cite{huang2021unignn} &
  \checkmark &
   &
   &
  Paul Häusner, Jens Sjölund \\
    \cmidrule(lr){2-7}
 &
   UniGIN \cite{huang2021unignn} &
  \checkmark &
   &
   &
  Kalyan Nadimpalli \\
    \cmidrule(lr){2-7}
 &
  DHGCN* \cite{wei2021dynamic} &
   &
   &
  \checkmark &
  Tatiana Malygina \\
 \toprule
\textbf{SC} &
  SCCONV \cite{bunch2020simplicial} &
   &
   &
  \checkmark &
  Abdelwahed Khamis, Ali Zia, Mohammed Hassanin  \\
    \cmidrule(lr){2-7}

 &
  SNN \cite{ebli2020simplicial} &
   &
  \checkmark &
   & Jens Agerberg, Georg Bökman, Pavlo Melnyk
   \\
     \cmidrule(lr){2-7}

 &
  SAN \cite{giusti2022simplicial} &
   &
  \checkmark &
   &
     Luca Scofano, Indro Spinelli, Simone Scardapane, Simone \\
   & & & & & Fiorellino, Olga Zaghen, Lev Telyatnikov, Claudio \\
   & & & & & Battiloro, Guillermo Bernardez \textcolor{blue}{(first place)}\\
  \cmidrule(lr){2-7}
 &
  SCA \cite{hajij2022simplicial} &
   &
   &
  \checkmark &
  Aiden Brent \textcolor{orange}{(hon. mention)}\\
    \cmidrule(lr){2-7}
 &
  Dist2Cycle \cite{keros2022dist2cycle} &
   &
  \checkmark &
   &
  Ali Zia \\
    \cmidrule(lr){2-7}
 &
  SCoNe \cite{roddenberryglaze2021principled} &
   &
  \checkmark &
   &
  1. Odin Hoff Gardaa \textcolor{red}{(second place)} \\
  & & & & & 2. Aiden Brent \\
    \cmidrule(lr){2-7}

 &
  SCNN \cite{yang2022simplicial} &
   &
  \checkmark &
   &
  Maosheng Yang, Lucia Testa \\
    \cmidrule(lr){2-7}

 &
  SCCNN \cite{yang2023convolutional} &
   &
  \checkmark &
   &
  1. Maosheng Yang, Lucia Testa \\
  & & & & & 2. Jens Agerberg, Georg Bökman,
Pavlo Melnyk \textcolor{orange}{(hon. mention)}\\
    \cmidrule(lr){2-7}

 &
  SCN \cite{yang2022efficient} &
   &
  \checkmark &
   &
  Yixiao Yue \\ \toprule
\textbf{CC}&
  CWN \cite{bodnar2021weisfeiler} &
   &
  \checkmark &
  \checkmark &
  Dmitrii Gavrilev, Gleb Bazhenov, Suraj Singh \textcolor{red}{(second place)}\\
    \cmidrule(lr){2-7}
 &
  CAN \cite{giusti2022cell} &
   &
   &
  \checkmark &
     1. Luca Scofano, Indro Spinelli, Simone Scardapane, Simone \\
   & & & & & Fiorellino, Olga Zaghen, Lev Telyatnikov, Claudio \\
   & & & & & Battiloro, Guillermo Bernardez \textcolor{blue}{(first place)}\\ 
   & & & & & 2. Abraham Rabinowitz \\ \toprule
\textbf{CCC}&
  HOAN \cite{hajij2022higher} &
   &
   \checkmark& 
  \checkmark &
  1. Rubén Ballester, Manuel Lecha, Sergio Escalera \textcolor{blue}{(first place)} \\
  & & & & & 2. Aiden Brent \textcolor{red}{(second place)}\\ \bottomrule \\
  \vspace{0.5cm}

\end{tabular}

\begin{minipage}{0.6\textwidth}

\caption{Model implementations submitted to the Topological Deep Learning Challenge. We organize original models according to domain: hypergraph (HG), simplicial (SC), cellular (CC), and combinatorial (CCC). Task level indicates the rank on which a prediction is made.}\end{minipage}
\label{submissions-table}
\end{table}

\afterpage{\clearpage
\bibliography{references}
\bibliographystyle{icml2023}
\printendnotes

}

\end{document}